# TYDI QA: A Benchmark for Information-Seeking Question Answering in *Ty*pologically *Di*verse Languages

**Jonathan H. Clark**♦♣  **Eunsol Choi**♠  **Michael Collins**♠  **Dan Garrette**♠
**Tom Kwiatkowski**♠  **Vitaly Nikolaev**♦♥  **Jennimaria Palomaki**♦♥

Google Research
tydiqa@google.com

## Abstract

Confidently making progress on multilingual modeling requires challenging, trustworthy evaluations. We present TYDI QA —a question answering dataset covering 11 typologically diverse languages with 204K question-answer pairs. The languages of TYDI QA are diverse with regard to their typology—the set of linguistic features each language expresses—such that we expect models performing well on this set to generalize across a large number of the world's languages. We present a quantitative analysis of the data quality and example-level qualitative linguistic analyses of observed language phenomena that would not be found in English-only corpora. To provide a realistic information-seeking task and avoid priming effects, questions are written by people who *want* to know the answer, but *don't* know the answer yet, and the data is collected directly in each language without the use of translation.

## 1 Introduction

When faced with a genuine information need, everyday users now benefit from the help of automatic question answering (QA) systems on a daily basis with high-quality systems integrated into search engines and digital assistants. Their questions are *information-seeking*—they *want* to know the answer, but *don't* know the answer yet. Recognizing the need to align research with the impact it will have on real users, the community has responded with datasets of information-seeking questions such as WikiQA (Yang et al., 2015), MS MARCO (Nguyen et al., 2016), QuAC (Choi et al., 2018), and the Natural Questions (NQ) (Kwiatkowski et al., 2019).

However, many people that might benefit from QA systems do not speak English. The languages of the world exhibit an astonishing breadth of linguistic phenomena used to express meaning; the World Atlas of Language Structures (Comrie and Gil, 2005; Dryer and Haspelmath, 2013) categorizes over 2600 languages[1] by 192 typological features including phenomena such as word order, reduplication, grammatical meanings encoded in morphosyntax, case markings, plurality systems, question marking, relativization, and many more. If our goal is to build models that can accurately represent all human languages, we must evaluate these models on data that exemplifies this variety.

In addition to these typological distinctions, modeling challenges arise due to differences in the availability of monolingual data, the availability of (expensive) parallel translation data, how standardized the writing system is variable spacing conventions (e.g. Thai), and more. With these needs in mind, we present the first public large-scale multilingual corpus of *information-seeking* question-answer pairs—using a simple-yet-novel data collection procedure that is model-free and translation-free. Our goals in doing so are:

1. to enable research progress toward building high-quality question answering systems in roughly the world's top 100 languages;[2] and
2. to encourage research on models that behave well across the linguistic phenomena and data scenarios of the world's languages.

We describe the typological features of TYDI QA's languages and provide glossed examples of some relevant phenomena drawn from the data to provide researchers with a sense of the challenges present in non-English text that their models will need to handle (Section 5). We also provide an open-source baseline model[3] and a public leaderboard[4] with a hidden test set to track community progress. We hope that enabling such intrinsic and extrinsic analyses on a challenging task will spark progress in multilingual modeling.

---

[1] Ethnologue catalogs over 7000 living languages.
[2] Despite only containing 11 languages, TYDI QA covers a large variety of linguistic phenomena and data scenarios.

The underlying data of a research study can have a strong influence on the conclusions that will be drawn: *Is QA solved? Do our models accurately represent a large variety of languages?* Attempting to answer these questions while experimenting on artificially easy datasets may result in overly optimistic conclusions that lead the research community to abandon potentially fruitful lines of work. We argue that TYDI QA will enable the community to reliably draw conclusions that are aligned with people's information-seeking needs while exercising systems' ability to handle a wide variety of language phenomena.

## 2 Task definition

TYDI QA presents a model with a question along with the content of a Wikipedia article, and requests that it make two predictions:

1. **Passage Selection Task:** Given a list of the passages in the article, return either (a) the index of the passage that answers the question or (b) NULL if no such passage exists.

2. **Minimal Answer Span Task:** Given the full text of an article, return one of (a) the start and end byte indices of the minimal span that completely answers the question; (b) YES or NO if the question requires a yes/no answer and we can draw a conclusion from the passage; (c) NULL if it is not possible to produce a minimal answer for this question.

Figure 1 shows an example question-answer pair. This formulation reflects that information-seeking users do not know where the answer to their question will come from, nor is it always obvious whether their question is even answerable.

## 3 Data collection procedure

**Question elicitation:** Human annotators are given short prompts consisting of the first 100 characters of Wikipedia articles and asked to write

---

[3]github.com/google-research-datasets/tydiqa
[4]ai.google.com/research/tydiqa

QUESTION: What are the types of matter?
ANSWER: ... Four states of matter are observable in everyday life: **solid, liquid, gas, and plasma**. Many other states are known to exist, such as glass or liquid crystal...

Figure 1: An English example from TYDI QA. The **answer passage** must be selected from a list of passages in a Wikipedia article while the **minimal answer** is some span of bytes in the article (bold). Many questions have no answer.

questions that (a) they are *actually interested in* knowing the answer to, and (b) that are *not answered by the prompt* (see Section 3.1 for the importance of unseen answers). The prompts are provided merely as inspiration to generate questions on a wide variety of topics; annotators are encouraged to ask questions that are only vaguely related to the prompt. For example, given the prompt *Apple is a fruit...*, an annotator might write *What disease did Steve Jobs die of?* We believe this stimulation of curiosity reflects how questions arise naturally: people encounter a stimulus such as a scene in a movie, a dog on the street, or an exhibit in a museum and their curiosity results in a question.

Our question elicitation process is similar to QuAC in that question writers see only a small snippet of Wikipedia content. However, QuAC annotators were requested to ask about a particular entity while TYDI QA annotators were encouraged to ask about anything interesting that came to mind, no matter how unrelated. This allows the question writers even more freedom to ask about topics that truly interest them, including topics not covered by the prompt article.

**Article retrieval:** A Wikipedia article[5] is then paired with each question by performing a Google search on the question text, restricted to the Wikipedia domain for each language, and selecting the top-ranked result. To enable future use cases, article text is drawn from an atomic Wikipedia snapshot of each language.[6]

**Answer labeling:** Finally, annotators are presented with the question/article pair and asked first to select the best **passage answer**—a paragraph[7] in the article that contains an answer—or else indicate that no answer is possible (or that no single passage is a satisfactory answer). If such a passage is found, annotators are asked to select, if possible, a **minimal answer**: a character span that is as short as possible while still forming a satisfactory answer to the question; ideally, these are 1–3 words long, but in some cases can span most of a sentence (e.g. for definitions such as *What is an atom?*). If the question is asking for a boolean answer, the annotator selects either YES or NO. If no such minimal answer is possible, then the annotators indicate this.

---

[5]We removed tables, long lists, and info boxes from the articles to focus the modeling challenge on multilingual text.
[6]Each snapshot corresponds to an Internet Archive URL.
[7]Or other roughly paragraph-like HTML element.

### 3.1 The importance of unseen answers

Our question writers *seek* information on a topic that they find interesting yet somewhat unfamiliar. When questions are formed without knowledge of the answer, the questions tend to contain (a) underspecification of questions such as *What is sugar made from?*—Did the asker intend a chemical formula or the plants it is derived from?—and (b) mismatches of the lexical choice and morphosyntax between the question and answer since the question writers are not cognitively primed to use the same words and grammatical constructions as some unseen answer. The resulting question-answer pairs avoid many typical artifacts of QA data creation such as high lexical overlap, which can be exploited by machine learning systems to artificially inflate task performance.[8]

We see this difference borne out in the leaderboards of datasets in each category: datasets where question writers saw the answer are mostly solved—for example, SQuAD (Rajpurkar et al., 2016, 2018) and CoQA (Reddy et al., 2019); datasets whose question writers did not see the answer text remain largely unsolved—for example, the Natural Questions (Kwiatkowski et al., 2019) and QuAC. Similarly, Lee et al. (2019) found that question answering datasets in which questions were written while annotators saw the answer text tend to be easily defeated by TF-IDF approaches that rely mostly on lexical overlap whereas datasets where question-writers did not know the answer benefited from more powerful models. Put another way, artificially easy datasets may favor overly simplistic models.

Unseen answers provide a natural mechanism for creating questions that are not answered by the text since many retrieved articles indeed do not contain an appropriate answer. In SQuAD 2.0 (Rajpurkar et al., 2018), unanswerable questions were artificially constructed.

---

[8]Compare these information-seeking questions with carefully-crafted reading comprehension or trivia questions that should have an unambiguous answer. There, expert question askers have a different purpose: to *validate* the knowledge of the potentially-expert question answerer.

### 3.2 Why not translate?

One approach to creating multilingual data is to translate an English corpus into other languages, as in XNLI (Conneau et al., 2018). However, the process of translation—including human translation—tends to introduce problematic artifacts to the output language such as preserving source-language word order as when translating from English to Czech (which allows flexible word order) or the use of more constrained language by translators (e.g. more formal). The result is that a corpus of so-called **Translationese** may be markedly different from purely native text (Lembersky et al., 2012; Volansky et al., 2013; Avner et al., 2014; Eetemadi and Toutanova, 2014; Rabinovich and Wintner, 2015; Wintner, 2016). Questions that originate in a different language may also differ in what is left underspecified or in what topics will be discussed. For example, in TYDI QA, one Bengali question asks *What does sapodilla taste like?*, referring to a fruit that is unlikely to be mentioned in an English corpus, presenting unique challenges for transfer learning. Each of these issues makes a translated corpus more English-like, potentially inflating the apparent gains of transfer-learning approaches.

Two recent multilingual QA datasets have used this approach. MLQA (Lewis et al., 2019) includes 12k SQuAD-like English QA instances; a subset of articles are matched to six target language articles via a multilingual model and the associated questions are translated. XQuAD (Artetxe et al., 2019) includes 1,190 QA instances from SQuAD 1.1, with both questions and articles translated into 10 languages.[9] Compared to TYDI QA, these datasets are vulnerable to Translationese while MLQA's use of a model-in-the-middle to match English answers to target language answers comes with some risks: (1) of selecting answers containing machine-translated Wikipedia content; and (2) of the dataset favoring models that are trained on the same parallel data or that use a similar multilingual model architecture.

---

[9]XQuAD translators see English questions and passages at the same time, priming them to use similar words.

## 3.3 Document-level reasoning

TYDI QA requires reasoning over lengthy articles (5K–30KB avg., Table 4) and a substantial portion of questions (46%–82%) cannot be answered by their article. This is consistent with the information-seeking scenario: the question asker does not wish to specify a small passage to scan for answers, nor is an answer guaranteed. In SQuAD-style datasets such as MLQA and XQuAD, the model is provided only a paragraph that always contains the answer. Full documents allow TYDI QA to embrace the natural ambiguity over correct answers, which is often correlated with difficult, interesting questions.

## 3.4 Quality control

To validate the quality of questions, we sampled questions from each annotator and verified with native speakers that the text was fluent.[10] We also verified that annotators were not asking questions answered by the prompts. We provided minimal guidance about acceptable questions, discouraging only categories such as opinions (e.g. *What is the best kind of gum?*) and conversational questions (e.g. *Who is your favorite football player?*).

Answer labeling required more training, particularly defining minimal answers. For example, should minimal answers include function words? Should minimal answers for definitions be full sentences? (Our guidelines specify no to both). Annotators performed a training task, requiring 90%+ to qualify. This training task was repeated throughout data collection to guard against annotators drifting off the task definition. We monitored inter-annotator agreement during data collection. For the dev and test sets,[11] a separate pool of annotators verified the questions and minimal answers to ensure that they are acceptable.[12]

## 4 Related Work

In addition to the various datasets discussed throughout Section 3, multilingual QA data has also been generated for very different tasks. For example, in XQA (Liu et al., 2019a) and XCMRC (Liu et al., 2019b), statements phrased syntactically as questions (*Did you know that ___ is the largest stringray?*) are given as prompts to retrieve a noun phrase from an article. Kenter et al. (2018) locate a span in a document that provides information on a certain property such as *location*.

Prior to these, several non-English multilingual question answering datasets have appeared, typically including one or two languages: These include DuReader (He et al., 2017) and DRCD (Shao et al., 2018) in Chinese, French/Japanese evaluation sets for SQuAD created via translation (Asai et al., 2018), Korean translations of SQuAD (Lee et al., 2018; Lim et al., 2019), a semi-automatic Italian translation of SQuAD (Croce et al., 2018), ARCD—an Arabic reading comprehension dataset (Mozannar et al., 2019), a Hindi-English parallel dataset in a SQuAD-like setting (Gupta et al., 2018), and a Chinese-English dataset focused on visual QA (Gao et al., 2015). The recent MLQA and XQuAD datasets also translate SQuAD in several languages (see Section 3.2). With the exception of DuReader, these sets also come with the same lexical overlap caveats as SQuAD.

Outside of QA, XNLI (Conneau et al., 2018) has gained popularity for natural language understanding. However, SNLI (Bowman et al., 2015) and MNLI (Williams et al., 2018) can be modeled surprisingly well while ignoring the presumably-critical premise (Poliak et al., 2018). While NLI stress tests have been created to mitigate these issues (Naik et al., 2018), constructing a representative NLI dataset remains an open area of research.

The question answering *format* encompasses a wide variety of tasks (Gardner et al., 2019) ranging from generating an answer word-by-word (Mitra, 2017) or finding an answer from within an entire corpus as in TREC (Voorhees and Tice, 2000) and DrQA (Chen et al., 2017).

Question answering can also be interpreted as an exercise in *verifying* the knowledge of experts by finding the answer to trivia questions that are carefully crafted by someone who already knows the answer such that exactly one answer is correct such as TriviaQA and Quizbowl/Jeopardy! questions (Ferrucci et al., 2010; Dunn et al., 2017; Joshi et al., 2017; Peskov et al., 2019); this information-verifying paradigm also describes reading comprehension datasets such as NewsQA (Trischler et al., 2017), SQuAD (Rajpurkar et al., 2016, 2018), CoQA (Reddy et al., 2019), and the multiple choice RACE (Lai et al., 2017). This paradigm has been taken even further by biasing the distribution of questions toward especially hard-to-model examples as in QAngaroo (Welbl et al., 2018), HotpotQA (Yang et al., 2018), and DROP (Dua et al., 2019). Others have focused exclusively on particular answer types such as boolean questions (Clark et al., 2019). Recent work has also sought to bridge the gap between dialog and QA, answering a series of questions in a conversational manner as in CoQA (Reddy et al., 2019) and QuAC (Choi et al., 2018).

---

[10] Small typos are acceptable as they are representative of how real users interact with QA.

[11] Except Finnish and Kiswahili.

[12] For questions, we accepted questions with minor typos or dialect, but rejected questions that were obviously non-native. For final-pass answer filtering, we rejected answers that were obviously incorrect, but accept answers that are plausible.

## 5 Typological diversity

Our primary criterion for including languages in this dataset is **typological diversity**—that is, the degree to which they express meanings using different linguistic devices, which we discuss below. In other words, we seek to include not just many *languages*, but many language *families*.

Furthermore, we select languages that have **diverse data characteristics** that are relevant to modeling. For example, some languages may have very little monolingual data. There are many languages with very little parallel translation data and for which there is little economic incentive to produce a large amount of expensive parallel data in the near future. Approaches that rely too heavily on the availability of high-quality machine translation will fail to generalize across the world's languages. For this reason, we select some languages that have parallel training data (e.g. Japanese, Arabic) and some that have very little parallel training data (e.g. Bengali, Kiswahili). Despite the much greater difficulties involved in collecting data in these languages, we expect that their diversity will allow researchers to make more reliable conclusions about how well their models will generalize across languages.

### 5.1 Discussion of Languages

We offer a comparative overview of linguistic features of the languages in TYDI QA in Table 1. To provide a glimpse into the linguistic phenomena that have been documented in the TYDI QA data, we discuss some of the most interesting features of each language below. These are by no means exhaustive, but rather intended to highlight the breadth of phenomena that this group of languages covers.

**Arabic:** Arabic is a Semitic language with short vowels indicated as typically-omitted diacritics. Arabic employs a root-pattern system: a sequence of consonants represents the root; letters vary *inside* the root to vary the meaning. Arabic relies on substantial affixation for inflectional and derivational word formation. Affixes also vary by grammatical number: singular, *dual* (two), and plural (Ryding, 2005). Clitics[13] are common (Attia, 2007).

**Bengali:** Bengali is a morphologically-rich language. Words may be complex due to inflection, affixation, compounding, reduplication, and the idiosyncrasies of the writing system including non-decomposable consonant conjuncts. (Thompson, 2010).

**Finnish:** Finnish is a Finno-Ugric language with rich inflectional and derivational suffixes. Word stems often alter due to morphophonological alternations (Karlsson, 2013). A typical Finnish noun has approximately 140 forms and a verb about 260 forms (Hakulinen et al., 2004).[14]

**Japanese:** Japanese is a mostly non-configurational[15] language in which particles are used to indicate grammatical roles though the verb typically occurs in the last position (Kaiser et al., 2013). Japanese uses 4 alphabets: kanji (ideograms shared with Chinese), hiragana (a phonetic alphabet for morphology and spelling), katakana (a phonetic alphabet for foreign words), and the Latin alphabet (for many new Western terms); all of these are in common usage and can be found in TYDI QA.

**Indonesian:** Indonesian is an Austronesian language characterized by reduplication of nouns, pronouns, adjectives, verbs, and numbers (Sneddon et al., 2012; Vania and Lopez, 2017), as well as prefixes, suffixes, infixes, and circumfixes.

**Kiswahili:** Kiswahili is a Bantu language with complex inflectional morphology. Unlike the majority of world languages, inflections, like number and person, are encoded in the prefix, not the suffix (Ashton, 1947). Noun modifiers show extensive agreement with the noun class (Mohamed,

---

[13]**Clitics** are affix-like linguistic elements that may carry grammatical or discourse-level meaning.

[14]Not counting forms derived through compounding or the addition of particle clitics.

[15]Among other linguistics features, 'non-configurational' languages exhibit generally free word order.

| LANGUAGE | LATIN SCRIPT[a] | WHITE SPACE TOKENS | SENTENCE BOUNDARIES | WORD FORMATION[b] | GENDER[c] | PRODROP |
|---|---|---|---|---|---|---|
| ENGLISH | + | + | + | + | +[d] | — |
| ARABIC | — | + | + | ++ | + | + |
| BENGALI | — | + | + | + | + | + |
| FINNISH | + | + | + | +++ | — | — |
| INDONESIAN | + | + | + | + | — | + |
| JAPANESE | — | — | + | + | — | + |
| KISWAHILI | + | + | + | +++ | —[e] | + |
| KOREAN | — | +[f] | + | +++ | — | + |
| RUSSIAN | + | + | + | ++ | + | + |
| TELUGU | — | + | + | +++ | + | + |
| THAI | — | — | — | + | + | + |

[a] '—' indicates **Latin script** is not the conventional writing system. Intermixing of Latin script should still be expected.
[b] We include inflectional and derivation phenomena in our notion of **word formation**.
[c] We limit the **gender** feature to sex-based gender systems associated with coreferential gendered personal pronouns.
[d] English has grammatical gender only in third person personal and possessive pronouns.
[e] Kiswahili has morphological noun classes (Corbett, 1991), but here we note sex-based gender systems.
[f] In Korean, tokens are often separated by whitespace, but prescriptive spacing conventions are commonly flouted.

Table 1: Typological features of the 11 languages in TYDI QA. We use $+$ to indicate that this phenomena occurs, $++$ to indicate that it occurs frequently, and $+++$ to indicate very frequently.

2001). Kiswahili is a pro-drop language[16] (Seidl and Dimitriadis, 1997; Wald, 1987). Most semantic relations that would be represented in English as prepositions are expressed in verbal morphology or by nouns (Wald, 1987).

**Korean:** Korean is an agglutinative, predicate-final language with a rich set of nominal and verbal suffixes and postpositions. Nominal particles express up to 15 cases—including the connective 'and'/'or'—and can be stacked in order of dominance from right to left. Verbal particles express a wide range of tense-aspect-mood, and include a devoted 'sentence-ender' for declarative, interrogative, imperative, etc. Korean also includes a rich system of honorifics. There is extensive discourse-level pro-drop (Sohn, 2001). The written system is a non-Latin featural alphabet arranged in syllabic blocks. White space is used in writing, but prescriptive conventions for spacing predicate-auxiliary compounds and semantically close noun-verb phrases are commonly flouted (Han and Ryu, 2005).

**Russian:** Russian is an Eastern Slavic language using the Cyrillic alphabet. An inflected language, it relies on case marking and agreement to represent grammatical roles. Russian uses singular, paucal,[17] and plural number. Substantial fusional[18] morphology (Comrie, 1989) is used along with three grammatical genders (Corbett, 1982), extensive prodrop (Bizzarri, 2015), and flexible word order (Bivon, 1971).

**Telugu:** Telugu is a Dravidian language. Orthographically, consonants are fully specified and vowels are expressed as diacritics if they differ from the default syllable vowel. Telugu is an agglutinating, suffixing language (Lisker, 1963; Krishnamurti, 2003). Nouns have 7-8 cases, singular/plural number, and three genders (feminine, masculine, neuter). An outstanding feature of Telugu is a productive process for forming transitives and causative forms (Krishnamurti, 1998).

**Thai:** Thai is an analytic language[19] despite very infrequent use of whitespace: Spacing in Thai is usually used to indicate the end of a sentence but may also indicate a phrase or clause break or appear before or after a number (Dānwiwat, 1987).

---

[16] Both the subject and the object can be dropped due to verbal inflection.

[17] **Paucal** number represents a few instances—between singular and plural. In Russian, paucal is used for quantities of 2, 3, 4, and many numerals ending in these digits.

[18] **Fusional** morphology expresses several grammatical categories in one unsegmentable element.

[19] An **analytic language** uses helper words rather than morphology to express grammatical relationships.

Q: Kuka keksi viiko-n-päivä-t ?
   who invented week-GEN-day-PL ?
   *Who invented the days of the week?*

A: Seitsen-päivä-inen viikko on
   seven-NOM-day-PL.ADJ week-NOM is
   todennäköisesti lähtöisin Babylonia-sta...
   likely origin Babylonia-ELA
   *The seven-day week is likely from Babylonia.*

Figure 2: Finnish example exhibiting compounding, inflection, and consonant gradation. In the question, *weekdays* is a compound. However, in the compound, *week* is inflected in the genitive case *-n* and the change of *kk* to *k* in the stem (a common morphophonological process in Finnish known as consonant gradation). The plural is marked on the head of the compound *day* by the plural suffix *-t*. But in the answer, *Week* is present as a standalone word in the nominative case (no overt case marking), but is modified by a compound adjective composed of *seven* and *days*.

Q: Как далеко Уран от
   how far Uranus-SG.NOM from
   Земл-и?
   Earth-SG.GEN?
   *How far is Uranus from Earth?*

A: Расстояние между Уран-ом
   distance between Uranus-SG.INSTR
   и Земл-ёй меняется от 2,6
   and Earth-SG.INSTR varies from 2,6
   до 3,15 млрд км...
   to 3,15 bln km...
   *The distance between Uranus and Earth fluctuates from 2.6 to 3.15 bln km...*

Figure 3: Russian example of morphological variation across question-answer pairs due to the difference in syntactic context: the entities are identical but have different representation, making simple string matching more difficult. The names of the planets are in the subject (Уран, Uranus-NOM) and object of the preposition (от земли, from Earth-GEN) context in the question. The relevant passage with the answer has the names of the planets in a coordinating phrase that is an object of a preposition (между Ураном и Землёй, between Uranus-INSTR and Earth-INSTR). Because the syntactic contexts are different, the names of the planets have different case marking.

Q: من هو موزارت ؟
   mn hw mwzArt ?
   *Who is **Muzart**?*
A: أماديوس موتسارت
   >mAdyws mwtsArt
   *...Amadeus **Mozart** ...*

Figure 4: Arabic example of inconsistent name spellings; both spellings are correct and refer to the same entity.

Q: ما هي الوان العلم العُماني ؟
   mA hy AlwAn AlElm AlEumAny ?
   *What are the colors of the **Omani** flag?*
A: العلم الوطني لسلطنة عمان
   AlElm AlwTny lslTnp EmAn
   *...the national flag of **Oman** ...*

Figure 5: Arabic example of selective diacritization. Note that the question contains diacritics (short vowels) to emphasize the pronunciation of *AlEumAny* (the specific entity intended) while the answer does not have diacritics in *EmAn*.

Q: متى ولد عبدالسلام بن محمد؟
   mtY wld EbdAlslAm bn mHmd ?
   *When was **AbdulSalam** bin Muhammad born?*
A: عبد السلام بن محمد بن أحمد
   Ebd AlslAm bn mHmd bn >Hmd
   *...**Abdul Salam** bin Muhammed bin Ahmed ...*

Figure 6: Arabic example of name de-spacing. The name appears as *AbdulSalam* in the question and *Abdul Salam* in the answer. This is potentially because of the visual break in the script between the two parts of the name. In manual orthography, the presence of the space would be nearly undetectable; its existence becomes an issue only in the digital realm.

Q: ما هي اول سيمفونية لبيتهوفن؟
   mA hy Awl symfwnyp lbythwfn ?
   *What is Beethoven's **first** symphony?*
A: السيمفونية الأولى لبيتهوفن
   Alsymfwnyp Al>wlY lbythwfn
   *...the **first** symphony for Beethoven ...*

Figure 7: Arabic example of gender variation of the word *first* (*Awl* vs *Al>wlY*) between the question and answer.

---

Additional glossed examples are available at ai.google.com/research/tydiqa.

## 5.2 A linguistic analysis

While the field of computational linguistics has remained informed by its roots in linguistics, practitioners often express a disconnect: descriptive linguists focus on fascinating complex phenomena, yet datasets that computational linguists encounter often do not contain such examples. TYDI QA is intended to help bridge this gap: we have identified and annotated examples from the data that exhibit linguistic phenomena that (a) are typically not found in English and (b) are potentially problematic for NLP models.

Figure 2 presents the interaction among three phenomena in a Finnish example, and Figure 3 shows an example of non-trivial word form changes due to inflection in Russian. Arabic also exemplifies many phenomena that are likely to challenge current models including spelling variation of names (Figure 4), selective diacritization of words (Figure 5), inconsistent use of whitespace (Figure 6), and gender variation (Figure 7).

These examples illustrate that the subtasks that are nearly trivial in English—such as string matching—can become complex for languages where morphophonological alternations and compounding cause dramatic variations in word forms.

## 6 A quantitative analysis

At a glance, TYDI QA consists of 204K examples: 166K are one-way annotated, to be used for training, and 37K are 3-way annotated, comprising the dev and test sets, for a total of 277K annotations (Table 4).

### 6.1 Question analysis

While we strongly suspect that the *relationship* between the question and answer is one of the best indicators of a QA dataset's difficulty, we also provide a comparison between the English question types found in TYDI QA and SQuAD in Table 2. Notably, TYDI QA displays a more balanced distribution of question words.[20]

### 6.2 Question-prompt analysis

We also evaluate how effectively the annotators followed the question elicitation protocol of Sec-tion 3. From a sample of 100 prompt-question pairs, we observed that all questions had 1–2 words of overlap with the prompt (typically an entity or word of interest) and none of the questions were answered by the prompt, as requested. Since these prompts are entirely discarded in the final dataset, the questions often have less lexical overlap with their answers than the prompts.

---

[20] For non-English languages, it is difficult to provide an intuitive analysis of question words across languages since question words can function differently depending on context.

| QUESTION WORD | TYDI QA | SQUAD |
|---|---|---|
| WHAT | 30% | 51% |
| HOW | 19% | 12% |
| WHEN | 14% | 8% |
| WHERE | 14% | 5% |
| (YES/NO) | 10% | <1% |
| WHO | 9% | 11% |
| WHICH | 3% | 5% |
| WHY | 1% | 2% |

Table 2: Distribution of question words in the English portion of the development data.

| NULL | PASSAGE ANSWER | MINIMAL ANSWER |
|---|---|---|
| 85% | 92% | 93% |

Table 3: Expert judgments of annotation accuracy. NULL indicates how often the annotation is correct given that an annotator marked a NULL answer. **Passage answer** and **minimal answer** indicate how often each is correct given the annotator marked an answer.

### 6.3 Data quality

In Table 3, we analyze the degree to which the annotations are *correct*.[21] Human experts[22] carefully judged a sample of 200 question-answer pairs from the dev set for Finnish and Russian. For each question, the expert indicates (1) whether or not each question has an answer within the article—the NULL column, (2) whether or not each of the three passage answer annotations is correct, and (3) whether the minimal answer is correct. We take these high accuracies as evidence that the quality of the dataset provides a useful and reliable

---

[21] We measure correctness instead of inter-annotator agreement since question may have multiple correct answers. For example, We have observed a yes/no question where both YES and NO were deemed correct. Aroyo and Welty (2015) discuss the pitfalls of over-constrained annotation guidelines in depth.

[22] Trained linguists with experience in NLP data collection.

| Language | Train (1-way) | Dev (3-way) | Test (3-way) | Avg. Question Tokens | Avg. Article Bytes | Avg. Answer Bytes | Avg. Passage Candidates | %With Passage Answer | %With Minimal Answer |
|---|---|---|---|---|---|---|---|---|---|
| (English) | 9,211 | 1031 | 1046 | 7.1 | 30K | 57 | 47 | 50% | 42% |
| Arabic | 23,092 | 1380 | 1421 | 5.8 | 14K | 114 | 34 | 76% | 69% |
| Bengali | 10,768 | 328 | 334 | 7.5 | 13K | 210 | 34 | 38% | 35% |
| Finnish | 15,285 | 2082 | 2065 | 4.9 | 19K | 74 | 35 | 49% | 41% |
| Indonesian | 14,952 | 1805 | 1809 | 5.6 | 11K | 91 | 32 | 38% | 34% |
| Japanese | 16,288 | 1709 | 1706 | — | 14K | 53 | 52 | 41% | 32% |
| Kiswahili | 17,613 | 2288 | 2278 | 6.8 | 5K | 39 | 35 | 24% | 22% |
| Korean | 10,981 | 1698 | 1722 | 5.1 | 12K | 67 | 67 | 26% | 22% |
| Russian | 12,803 | 1625 | 1637 | 6.5 | 27K | 106 | 74 | 64% | 51% |
| Telugu | 24,558 | 2479 | 2530 | 5.2 | 7K | 279 | 32 | 28% | 27% |
| Thai | 11,365 | 2245 | 2203 | — | 14K | 171 | 38 | 54% | 43% |
| TOTAL | 166,916 | 18,670 | 18,751 | | | | | | |

Table 4: Data Statistics. Data properties vary depends on languages, as documents on Wikipedia differ significantly and annotators don't overlap between languages. We include a small amount of English data for debugging purposes, though we do not include English in macro-averaged results, nor in the leaderboard competition. Note that a single character may occupy several bytes in non-Latin alphabets.

signal for the assessment of multilingual question answering models.

Looking into these error patterns, we see that the NULL-related errors are entirely false positives (failing to find answers that exist), which would largely be mitigated by having 3 answer annotations. Such errors occur in a variety of article lengths from under 1000 words through large 3000-word articles. Therefore, we cannot attribute NULL errors to long articles alone, but we should consider alternative causes such as some question-answer matching being more difficult or subtle.

For minimal answers, errors occur for a large variety of reasons. One error category is when multiple dates seem plausible but only one is correct. One Russian question reads *When did Valentino Rossi win the first title?*. Two annotators correctly selected 1997 while one selected 2001, which was visually prominent in a large list of years.

## 7 Evaluation

### 7.1 Evaluation measures

We now turn from analyzing the quality of the data itself toward how to evaluate question answering systems using the data. The TYDI QA task's primary evaluation measure is F1, a harmonic mean of precision and recall, each of which is calculated over the examples within a language. However, certain nuances do arise for our task.

**NULL handling:** TYDI QA is an imbalanced dataset in terms of whether or not each question has an answer due to differing amounts of content in each language on Wikipedia. However, it is undesirable if a strategy such as always predicting NULL can produce artificially inflated results—this would indeed be the case if we were to give credit to a system producing NULL if *any* of the three annotators selected a NULL answer. Therefore, we first use a threshold to select a **NULL consensus** for each evaluation example: at least 2 of the 3 annotators must select an answer for the consensus to be non-NULL. The NULL consensus for the given task (passage answer, minimal answer) must be NULL in order for a system to receive credit (see below) for a NULL prediction.

**Passage selection task:** For questions having a NULL consensus (see above), credit is given for matching any of the passage indices selected by annotators.[23] An example counts toward the denominator of recall if it has a non-NULL consensus, and toward the denominator of precision if the model predicted a non-NULL answer.

**Minimal span task:** For each example, given the question and text of an article, a system must predict NULL, YES, NO, or a contiguous span of bytes that constitutes the answer. For span answers, we treat this collection of byte index pairs as a set and compute an example-wise F1 score between each annotator's minimal answer and the model's minimal answer, with partial credit assigned when spans are partially overlapping; the maximum is returned as the score for each example. For a YES/NO answers, credit is given (a score of 1.0), if any of the annotators indicated such as a correct answer. The NULL consensus must be non-NULL in order to receive credit for a non-NULL answer.

---
[23]By matching *any* passage, we effectively take the max over examples, consistent with the minimal span task.

|           | Train Size | Passage Answer F1 (P/R) | | | Minimal Answer Span F1 (P/R) | |
|-----------|-----------|---------------|-----------|--------------|----------|--------------|
|           |           | First passage | mBERT     | Lesser Human | mBERT    | Lesser Human |
| (English) | 9,211     | 32.9 (28.4/39.1) | 62.5 (62.6/62.5) | 69.4 (63.4/77.6) | 44.0 (52.9/37.8) | 54.4 (52.9/56.5) |
| Arabic    | 23,092    | 64.7 (59.2/71.3) | 81.7 (85.7/78.1) | 85.4 (82.1/89.0) | 69.3 (74.9/64.5) | 73.5 (73.6/73.5) |
| Bengali   | 10,768    | 21.4 (15.5/34.6) | 60.3 (61.4/59.5) | 85.5 (81.6/89.7) | 47.7 (50.7/45.3) | 79.1 (78.6/79.7) |
| Finnish   | 15,285    | 35.4 (28.4/47.1) | 60.8 (58.7/63.0) | 76.3 (69.8/84.2) | 48.0 (56.7/41.8) | 65.3 (61.8/69.4) |
| Indonesian| 14,952    | 32.6 (23.8/51.7) | 61.4 (57.2/66.7) | 78.6 (72.7/85.6) | 51.3 (54.5/48.8) | 71.1 (68.7/73.7) |
| Japanese  | 16,288    | 19.4 (14.8/28.0) | 40.6 (42.2/39.5) | 65.1 (57.8/74.8) | 30.4 (42.1/23.9) | 53.3 (51.8/55.2) |
| Kiswahili | 17,613    | 20.3 (13.4/42.0) | 60.2 (58.4/62.3) | 76.8 (70.1/85.0) | 49.7 (55.2/45.4) | 67.4 (63.4/72.1) |
| Korean    | 10,981    | 19.9 (13.1/41.5) | 56.8 (58.7/55.3) | 72.9 (66.3/82.4) | 40.1 (45.2/36.2) | 56.7 (56.3/58.6) |
| Russian   | 12,803    | 30.0 (25.5/36.4) | 63.2 (65.3/61.2) | 87.2 (84.4/90.2) | 45.8 (51.7/41.2) | 76.0 (82.0/70.8) |
| Telugu    | 24,558    | 23.3 (15.1/50.9) | 81.3 (81.7/80.9) | 95.0 (93.3/96.8) | 74.3 (77.7/71.3) | 93.3 (91.6/95.2) |
| Thai      | 11,365    | 34.7 (27.8/46.4) | 64.7 (61.8/68.0) | 76.1 (69.9/84.3) | 48.3 (54.3/43.7) | 65.6 (63.9/67.9) |
| **OVERALL** | 166,916 | 30.2 (23.6/45.0) | 63.1 (57.0/59.1) | 79.9 (84.4/74.5) | 50.5 (41.3/35.3) | 70.1 (70.8/62.4) |

Table 5: Quality on the TYDI QA primary tasks (passage answer and minimal answer) using: a naïve first-passage baseline, the open-source multilingual BERT model (mBERT), and a human predictor (Section 7.3). F1, precision, and recall measurements (Section 7.1) are averaged over four fine-tuning replicas for mBERT.

**Macro-averaging:** First, the scores for each example are averaged within a language; we then average over all *non-English* languages to obtain a final F1 score. Measurements on English are treated as a useful means of debugging rather than a goal of the TYDI QA task as there is already plenty of coverage for English evaluation in existing datasets.

### 7.2 An estimate of human performance

In this section, we consider two idealized methods for estimating human performance before settling on a widely used pragmatic method.

**A fair contest:** As a thought experiment, consider framing evaluation as "What is the likelihood that a correct answer is accepted as correct?" Trivia competitions and game shows take this approach as they are verifying the expertise of human answerers. One could exhaustively enumerate all correct passage answers; given several annotations of high accuracy, we would quickly obtain high recall. This approach is advocated in Boyd-Graber (2019).

**A game with preferred answers:** If our goal is to provide users with the answers that they *prefer*. If annotators correctly choose these preferred answers, we expect our multi-way annotated data to contain a distribution peaked around these preferred answers. The optimal strategy for players is then to predict those answers, which are both preferred by users and more likely to be in the evaluation dataset. We would expect a large pool of human annotators or a well-optimized machine learning system to learn this distribution. For example, the Natural Questions (Kwiatkowski et al., 2019) uses a 25-way annotations to construct a super-annotator, increasing the estimate of human performance by around 15 points F1.

**A lesser estimate of human performance:** Unfortunately, finding a very large pool of annotators for 11 languages would be prohibitively expensive. Instead, we provide a more pessimistic estimate of human performance by holding out one human annotation as a prediction and evaluating it against the other two annotations; we use bootstrap resampling to repeat this procedure for all possible combinations of 1 vs. 2 annotators. This corresponds to the human evaluation methodology for SQuAD with the addition of bootstrapping to reduce variance. In Table 5, we show this estimate of human performance. In cases where annotators disagree, this estimate will degrade, which may lead to an underestimate of human performance since in reality multiple answers could be correct. At first glance, these F1 scores may appear low compared to simpler tasks such as SQuAD, yet a *single* human prediction on the Natural Questions short answer task (similar to the TYDI QA minimal answer task), scores only 57 F1 even with the advantage of evaluating against five annotations rather than just two and training on 30X more English training data.

### 7.3 Primary tasks: Baseline results

To provide an estimate of the difficulty of this dataset for well-studied state-of-the-art models, we present results for a baseline that uses the most recently released multilingual BERT (mBERT)[24] (Devlin et al., 2019) in a setup similar to Alberti et al. (2019), in which all languages are trained jointly in a single model (Table 5). Additionally, as a naïve, untrained baseline, we include the results of a system that always predicts the first passage, since the first paragraph of a Wikipedia article often summarizes its most important facts. Across all languages, we see a large gap between mBERT and a lesser estimate of human performance (Section 7.2).

**Can we compare scores across languages?** Unfortunately, no. Each language has its own unique set of questions, varying quality and amount of Wikipedia content, quality of annotators, and other variables. We believe it is best to directly engage with these issues; avoiding these phenomena may hide important aspects of the problem space associated with these languages.

## 8 Gold passage: A simplified task

Up to this point, we have discussed the primary tasks of Passage Selection (SELECTP) and Minimal Answer Span (MINSPAN). In this section, we describe a simplified Gold Passage (GOLDP) task, which is more similar to existing reading comprehension datasets, with two goals in mind: (1) more directly comparing with prior work, and (2) providing a simplified way for researchers to use TYDI QA by providing compatibility with existing code for SQuAD, XQuAD, and MLQA.

---
[24] github.com/google-research/bert

|            | TYDIQA-GOLDP | MLQA | XQuAD |
|------------|--------------|------|-------|
| (English)  | 0.38         | 0.91 | 1.52  |
| Arabic     | 0.26         | 0.61 | 1.29  |
| Bengali    | 0.29         | —    | —     |
| Finnish    | 0.23         | —    | —     |
| Indonesian | 0.41         | —    | —     |
| Kiswahili  | 0.31         | —    | —     |
| Korean     | 0.19         | —    | —     |
| Russian    | 0.16         | —    | 1.13  |
| Telugu     | 0.13         | —    | —     |

Table 6: Lexical overlap statistics for TYDIQA-GOLDP, MLQA, and XQuAD showing the average number of tokens in common between the question and a 200-character window around the answer span. As expected, we observe substantially lower lexical overlap in TYDI QA.

Toward these goals, the Gold Passage task differs from the primary tasks in several ways:

- only the gold answer passage is provided rather than the entire Wikipedia article;
- unanswerable questions have been discarded, similar to MLQA and XQuAD;
- we evaluate with the SQuAD 1.1 metrics like XQuAD; and
- Thai and Japanese are removed since the lack of whitespace breaks some existing tools.

To better estimate human performance, only passages having 2+ annotations are retained. Of these annotations, one is withheld as a human prediction and the remainder are used as the gold set.

### 8.1 Gold passage lexical overlap

In Section 3, we argued that unseen answers and no translation should lead to a more complex, subtle relationship between the resulting questions and answers. We measure this directly in Table 6, showing the average number of tokens in common between the question and a 200-character window around the answer span, excluding the top 100 most frequent tokens, which tend to be non-content words. For all languages, we see a substantially lower lexical overlap in TYDI QA as compared to MLQA and XQuAD, corpora whose generation procedures involve seen answers and translation; we also see overall lower lexical overlap in non-English languages. We take this as evidence of a more complex relationship between questions and answers in TYDI QA.

|  | TYDIQA-GOLDP | SQuAD Zero Shot | Human |
|---|---|---|---|
| (English) | (76.8) | (73.4) | (84.2) |
| Arabic | 81.7 | 60.3 | 85.8 |
| Bengali | 75.4 | 57.3 | 94.8 |
| Finnish | 79.4 | 56.2 | 87.0 |
| Indonesian | 84.8 | 60.8 | 92.0 |
| Kiswahili | 81.9 | 52.9 | 92.0 |
| Korean | 69.2 | 50.0 | 82.0 |
| Russian | 76.2 | 64.4 | 96.3 |
| Telugu | 83.3 | 49.3 | 97.1 |
| OVERALL | 79.0 | 56.4 | 90.9 |

Table 7: F1 scores for the simplified TYDIQA-GOLDP task v1.1. *Left:* Fine tuned and evaluated on the TYDIQA-GOLDP set. *Middle:* Fine tuned on SQuAD v1.1 and evaluated on the TYDIQA-GOLDP dev set, following the XQuAD zero-shot setting. *Right:* Estimate of human performance on TYDIQA-GOLDP. Models are averaged over 5 fine tunings.

### 8.2 Gold passage results

In Table 7, we show the results of two experiments on this secondary Gold Passage task. First, we fine tune mBERT jointly on all languages of the TYDI QA gold passage training data and evaluate on its dev set. Despite lacking several of the core challenges of TYDI QA (e.g. no long articles, no unanswerable questions), F1 scores remain low, leaving headroom for future improvement.

Second, we fine tune on the 100k English-only SQuAD 1.1 training set and evaluate on the full TYDI QA gold passage dev set, following the XQuAD evaluation zero-shot setting. We again observe very low F1 scores. These are similar to, though somewhat lower than, the F1 scores observed in the XQuAD zero-shot setting of Artetxe et al. (2019). Strikingly, even the English performance is significantly lower, demonstrating that the style of question-answer pairs in SQuAD have very limited value in training a model for TYDI QA-style questions, despite the much larger volume of English questions in SQuAD.

## 9 Recommendations and future work

We foresee several research directions where this data will allow the research community to push new boundaries, including:

- studying the interaction between morphology and question-answer matching;
- evaluating the effectiveness of transfer learning, both for languages where parallel data is and is not available;
- the usefulness of machine translation in question answering for data augmentation and as a runtime component, given varying data scenarios and linguistic challenges;[25]
- studying zero-shot QA by explicitly not training on a subset of the provided languages.

We also believe that a deeper understanding of the data itself will be key and we encourage further linguistic analyses of the data. Such insights will help us understand what modeling techniques will be better-suited to tackling the full variety of phenomena observed in the world's languages.

We recognize that no single effort will be sufficient to cover the world's languages, and so we invite others to create compatible datasets for other languages; the universal dependency treebank (Nivre et al., 2016) now has over 70 languages, demonstrating what the community is capable of with broad effort.[26]

Finally, we note that the content required to answer questions often has simply not been written down in many languages. For these languages, we are paradoxically faced with the prospect that cross-language answer retrieval and translation are necessary, yet low-resource languages will also lack (and will likely continue to lack) the parallel data needed for trustworthy translation systems.

## 10 Conclusion

Confidently making progress on multilingual models requires challenging, trustworthy evaluations. We have argued that question answering is well-suited for this purpose and that by targeting a typologically diverse set of languages, progress on the TYDI QA dataset is more likely to generalize on the breadth of linguistic phenomena found throughout the world's languages. By avoiding data collection procedures reliant on translation and multilingual modeling, we greatly mitigate the risk of sampling bias. We look forward to the many ways the research community finds to improve the quality of multilingual models.

---

[25] Because we believe MT may be a fruitful research direction for TYDI QA, we do not release any automatic translations. In the past, this seems to have stymied innovation around translation as applied to multilingual datasets.

[26] We will happily share our annotation protocol on request.

## Acknowledgements

The authors wish to thank Chris Dyer, Daphne Luong, Dipanjan Das, Emily Pitler, Jacob Devlin, Jason Baldridge, Jordan Boyd-Graber, Kenton Lee, Kristina Toutanova, Mohammed Attia, Slav Petrov, and Waleed Ammar for their support, help analyzing data, and many insightful discussions about this work. We also thank Fadi Biadsy, Geeta Madhavi Kala, Iftekhar Naim, Maftuhah Ismail, Rola Najem, Taku Kudo, and Takaki Makino for their help in proofing the data for quality. We acknowledge Ashwin Kakarla and Karen Yee for support in data collection for this project.